%% file: main.tex
\documentclass{article} 
\usepackage{iclr2021_conference,times}

\input{math_commands.tex}

\usepackage{hyperref}
\usepackage{url}
\usepackage{todonotes}
\usepackage{booktabs}
\usepackage{graphicx}

\usepackage{xcolor}
\definecolor{darkred}{rgb}{.7,0,0}
\definecolor{darkgreen}{rgb}{0,.5,0}
\definecolor{darkblue}{rgb}{0,0,.8}
\definecolor{darkcyan}{rgb}{0,0.6,.6}
\definecolor{darkorange}{rgb}{.8,.4,0}
\definecolor{gray}{rgb}{.4,.4,.4}

\newcommand{\needcite}[1]{\textcolor{darkorange}{\emph{[citation needed\ifthenelse{\equal{#1}{}}{}{: #1}]}}}

\usepackage{algorithm}
\usepackage{subcaption}
\usepackage{listings} 

\usepackage{letltxmacro}
\newcommand*{\SavedLstInline}{}
\LetLtxMacro\SavedLstInline\lstinline
\DeclareRobustCommand*{\lstinline}{%
  \ifmmode
    \let\SavedBGroup\bgroup
    \def\bgroup{%
      \let\bgroup\SavedBGroup
      \hbox\bgroup
    }%
  \fi
  \SavedLstInline
}
\newcommand{\code}[1]{\lstinline[mathescape=true]{#1}}
\title{Training a First-Order Theorem Prover from Synthetic Data}

\author{%
  Vlad Firoiu \\
  DeepMind \\
  \texttt{vladfi@google.com} \\
  \And
  Eser Ayg\"un \\
  DeepMind \\
  \texttt{eser@google.com} \\
  \And Ankit Anand \\
  DeepMind \\
  \texttt{anandank@google.com} \\
  \And Zafarali Ahmed \\
  DeepMind \\
  \texttt{zaf@google.com} \\
  \And Xavier Glorot \\
  DeepMind \\
  \texttt{glorotx@google.com} \\
  \And Laurent Orseau \\
  DeepMind \\
  \texttt{lorseau@google.com} \\
  \And Lei Zhang \\
  DeepMind \\
  \texttt{lmzhang@google.com} \\
  \And Doina Precup \\
  DeepMind \\
  \texttt{doinap@google.com} \\
  \And Shibl Mourad \\
  DeepMind \\
  \texttt{shibl@google.com} \\
}

\arXiv  
\begin{document}

\maketitle

\begin{abstract}
A major challenge in applying machine learning to automated theorem proving is the scarcity of training data, which is a key ingredient in training successful deep learning models. To tackle this problem, we propose an approach that relies on training purely with synthetically generated theorems, without any human data aside from axioms. We use these theorems to train a neurally-guided saturation-based prover. Our neural prover outperforms the state-of-the-art E-prover on this synthetic data in both time and search steps, and shows significant transfer to the unseen human-written theorems from the TPTP library, where it solves 72\% of first-order problems without equality.
\end{abstract}

\input{introduction.tex}

\input{proposer.tex}
\input{prover.tex}
\input{results.tex}
\input{discussion.tex}

\bibliography{main}
\bibliographystyle{iclr2021_conference}

\appendix
\input{appendix.tex}

\end{document}

%% file: math_commands.tex
\usepackage{amsmath,amsfonts,bm}

\def\eqref#1{equation~\ref{#1}}

\def\1{\bm{1}}

\DeclareMathAlphabet{\mathsfit}{\encodingdefault}{\sfdefault}{m}{sl}
\SetMathAlphabet{\mathsfit}{bold}{\encodingdefault}{\sfdefault}{bx}{n}



%% file: introduction.tex
\section{Introduction}

Most work applying machine learning to theorem proving takes the following approach:
1) pick a dataset of formalized mathematics, such as Mizar or Metamath, or the standard library of a major proof assistant such as HOL-Light or Coq;
2) split the dataset into train and test;
3) use imitation learning or reinforcement learning on the training set to learn a policy;
and finally
4) evaluate the policy on the test set 
(\citet{loos&al17}, 
\citet{bansal2019holist}, 
\citet{yang&deng19}, 
\citet{han2021proof}, 
\citet{polu2020generative}) 
.
Such methods are fundamentally limited by the size of the training set, particularly when relying on deep neural networks \citep{kaplan2020scaling}. Unfortunately, unlike in computer vision and natural language processing, theorem proving datasets are comparatively tiny. Lean's mathlib, for example, measures only 22MB, 25000x smaller than the text data used in a powerful language model like GPT-3 \citep{brown2020language}. Due to the significant time and expertise required to write formalized mathematics, we believe an alternative path to obtaining data is required.

In this work we consider synthetically generating problems in order to train a neural theorem prover. We focus on ten first-order domains (without equality) from the TPTP problem library \citep{sutcliffe17}. Each domain has an \emph{axiom set} and \emph{problem set}. Our problem generator, the \emph{forward proposer}, randomly generates theorems given an axiom set. These theorems are then used to train a first-order resolution prover via a type of search distillation similar to AlphaZero~\citep{silver2018general}. This is a powerful training technique, allowing us to surpass the state-of-the-art E-prover~\citep{schulz&al19} on the synthetic training distribution.

Unlike in \citet{wang&deng20}, who train a theorem generator based on human-written theorems from Metamath, neither proposer nor prover see any TPTP problems until test time. Nevertheless, we see significant transfer from the synthetic distribution to TPTP test problems, with a jump from 42\% for our basic non-learning prover to 72\% for our best trained prover.

%% file: proposer.tex
\section{The Forward Proposer}

We obtain our training theorems using a simple but synthetic theorem generator, which we call the \emph{forward proposer} or \texttt{FwdP}. It starts from the clauses corresponding to the axioms of a specific domain of mathematics (e.g., geometry) and uses the resolution calculus \citep{fitting12} to infer new clauses. The forward proposer generates new clauses in this fashion for a certain number of steps, and uses the final clause as the conclusion of the theorem: Axioms $\rightarrow$ Clause. As the process starts with the axioms and the clauses are generated by logical inferences, this conjecture is guaranteed to be a valid theorem in the given domain.

To force the proposer towards clauses with deeper proof trees, at each step (except the first) we only allow inferences that involve the last generated clause. This is known as \emph{linear resolution}, and despite this restriction it is known to be as powerful as full resolution~\citep{fitting12}. See Appendix~\ref{appendix:fwdp-limitations} for a discussion of the forward proposer's limitations.

The simplest forward proposer samples linear resolutions uniformly at random at each step. However, this has a tendency to generate exponentially large clauses, which do not make for interesting theorems. So, we bias the resolutions towards smaller clauses using a softmax distribution based on the clause sizes, as measured by number of symbols.

For a given domain, the forward proposer is parameterized by two quantities: the number of forward steps $N$, and the softmax temperature $T$. Ultimately, we want to select these parameters in order to maximize transfer from synthetic to real data. However, as this is expensive to measure, and since we wish to be robust to unknown test problems, we used three proxy measures as criteria: size, difficulty, and diversity. Intuitively, smaller problems are more interesting by an Occam's Razor argument, and thus more likely similar to the test problems. It is important to train on difficult problems to be ready for difficult problems at test time. Finally, diversity is important for machine learning to generalize from train to test; insufficient diversity results in overfitting.

Thus, for each domain we performed a grid search over both $N$ and $T$ and chose the setting which maximized difficulty for the first-order E-prover~\citep{schulz&al19} while capping the mean generated clause size to 64.
We ensured diversity by sampling one million theorems per parameter setting, and requiring that there be at least 500K unique theorems among them; in practice this did not disqualify any relevant parameter settings. For more details see Appendix~\ref{appendix:fwdp-settings}.

%% file: prover.tex
\section{A Neural Resolution Prover}
\label{section:prover}

Our prover uses the first-order resolution calculus to find refutation proofs, similar to well-known automated theorem provers such as E and Vampire~\citep{kovacs2013first}. We use a simple implementation of the ``given-clause'' algorithm~\citep{mccune&wos97}, and do not support superposition or first-order equality. Briefly, the given-clause algorithm begins with a set of clauses to refute, typically the axioms and negated conjecture. Clauses are added to this ``active'' set one-by-one from the available inferences (resolutions and factorings) until falsehood (the empty clause) is derived, or time runs out. The selection of the ``given'' clause at each step is done by a cost function, which independently scores each available inference. Inference costs are only computed once, when the inference is made, and are not updated based on changing proof state. Because the number of available inferences can grow very rapidly, this loss in expressiveness comes with major efficiency gains, particularly when using expensive neural cost functions. For more a more detailed explanation of our prover and its features, see Appendix~\ref{appendix:saturation}.

Similarly to \citet{loos&al17}, we use a neural network cost function which is trained to predict whether a given clause will appear in the proof. However, rather than learning from a fixed dataset of proof attempts, we train on proofs generated by the neural prover itself. Similarly to online reinforcement learning, as the network improves, so does the data distribution.

More concretely, for each successful proof search, we determine which clauses were part of the found proof and use them as positive examples for the network. We then sample an equal number of negative examples uniformly from the remaining clauses which did not appear in the proof. Surprisingly, more sophisticated negative sampling methods did not help. Unsuccessful proof attempts are discarded and do not contribute to training.

\subsection{Architectures and Representations}

In this work we explored two different neural architectures for the cost function: 1) a multi-layer perceptron (MLP) on top of a fixed set of aggregated clause statistics, and 2) a Transformer~\citep{vaswani&al17} on top of a graphical representation of the clause with spectral features. Both models take as input the theorem to be proven and a target clause, and output a single logit representing the probability that the target clause will be used in the proof of the theorem. The theorem itself is represented by the initial clauses, which are either axioms or negated conjecture clauses. Since the axioms are the same for all theorems, we only expose the negated conjecture clauses to the model. As mentioned above, the cost function does not see the rest of the (typically very large) proof state.

The MLP model needs a fixed-size representation of the target and negated conjecture clauses. For each clause we use a small set of quantities such as the number of variables or literals in the clause, and aggregate over the negated conjecture clauses independently from the target. See Appendix~\ref{appendix:architectures} for full details.

The Transformer uses a more complete representation based on a graphical representation of the input. Each clause is transformed into its syntax tree, and the nodes are augmented with both type information (clause, literal, atom, or variable) and a vector hash of the symbol text. This way, each functor and predicate is given a stable feature vector. In sequence modeling, each input token to a transformer is given a \emph{positional} encoding representing its position in the sequence. Analogously, we give each node a \emph{spectral} encoding representing its position in the graph; this is given by the eigenvectors of the Laplacian matrix of the graph \citep{dwivedi2021generalization}.

%% file: results.tex
\section{Experiments and Results}

In our experiments, we used axiom sets and problems available in the TPTP library. These axioms come separated according to the different domains of mathematics and reasoning. We formed axiom sets by grouping axiom files that occur together in the same problems. We then filtered out any axiom sets with more than 1000 axiom clauses or less than ten associated theorems. We also excluded any domains that required first-order equality. This left us with ten axiom sets, covering  field theory (FLD), geometry (GEO), number theory (NUM), group theory (GRP), set theory (SET) and knowledge representation (KRS).  See Table~\ref{table:datasets} in the appendix for the numbers of axioms and TPTP theorems in each dataset.

For each domain, we trained an MLP and Transformer prover on forward proposer problems using the appears-in-proof prediction procedure described in Section~\ref{section:prover}. We then evaluated on both a random sample of 1000 forward proposer problems and the associated TPTP problem set. For comparison, we used two baselines: 1) a ``basic'' version of our prover with a simple (not learned) cost function based on clause size, and 2) E-prover version 2.5 with the \texttt{--auto} flag. All four provers were given 5 minutes per problem. See Appendix~\ref{appendix:experiments} for full experimental details.

\begin{figure}[h]
\centering
\begin{minipage}{.5\textwidth}
    \centering
    \includegraphics[width=\textwidth]{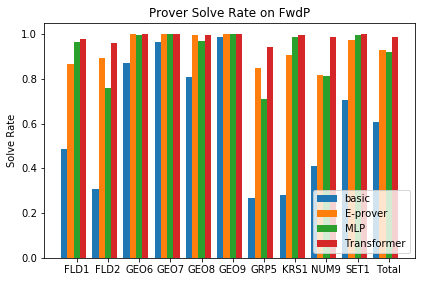}
\end{minipage}%
\begin{minipage}{.5\textwidth}
    \centering
    \includegraphics[width=\textwidth]{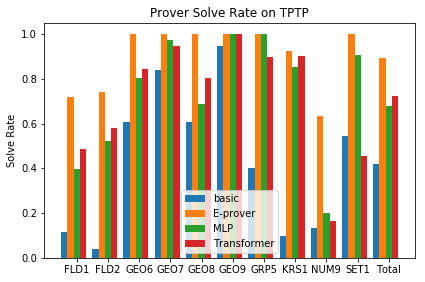}
\end{minipage}
\caption{Prover solve rates on FwdP and TPTP datasets.}
\label{fig:results}
\end{figure}

\begin{figure}[h]
\centering
\begin{minipage}{.5\textwidth}
    \centering
    \includegraphics[width=\textwidth]{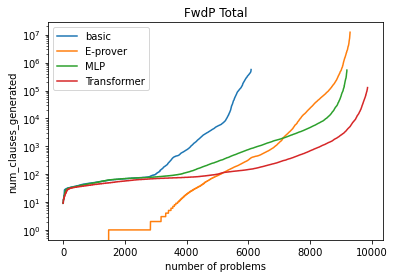}
\end{minipage}%
\begin{minipage}{.5\textwidth}
    \centering
    \includegraphics[width=\textwidth]{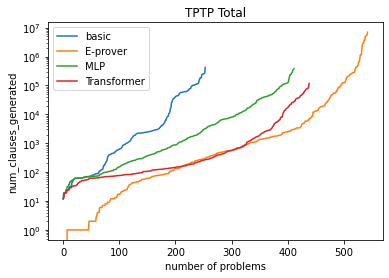}
\end{minipage}
\caption{Survival plots showing search steps (number of clauses generated) per problem versus number of problems solved. Results are aggregated across all problem domains. The transformer model consistently requires an order of magnitude fewer search steps than the MLP.}
\label{fig:survival}
\end{figure}

Overall, we found that the Transformer consistently outperformed the other provers on the synthetic problem distribution, in both problems solved and search steps (Figure~\ref{fig:survival}, left). Even the MLP nearly tied the state-of-the-art E-prover in terms of problems solved, and was significantly more efficient in search steps. These results validate the power of machine learning in the presence of plentiful data.

As expected, TPTP proved much harder for our neural provers, with results somewhat mixed between the MLP and Transformer depending on axiom set. However, we find it mildly surprising that the Transformer's greater capacity does not necessarily lead to overfitting on the significantly different training distribution; instead, the relative performance difference between the two is similar on FwdP and TPTP totals, in both search steps and total number of theorems proven. This indicates that genuinely useful proving heuristics are being learned from even the simple forward proposer's data.

%% file: discussion.tex
\section{Discussion}

The results presented here lend some credence to the hypothesis that synthetic problems can be used to train a theorem prover. However, it is clear that more sophisticated proposers are necessary going forward. For one, the transformer prover is close to 100\% on most synthetic domains, and so has little left to learn from the current Forward Proposer.

There have been various attempts at automated task generation outside of theorem proving 
(\citet{racaniere2020automated}, 
\citet{sukhbaatar2018intrinsic}, 
\citet{dennis2021emergent}, 
\citet{forestier2020intrinsically}).
Many of these are applicable, with some modifications, to theorem generation. For example, a policy or cost function could be trained to optimize a mixture of difficulty (as estimated by a judge) and clause size. Diversity could be maintained by injecting noise, co-training with a prover (so that the notion of difficulty is non-stationary), or by being measured with a generative model and optimized for.

There is a limit, however, to simply optimizing for size, difficulty, and diversity. Indeed, mathematics is full of uninteresting problems that meet these criteria, such as cryptographic instances of prime factorization or hash function inversion. What we would ultimately want is for a proposer to find \emph{interesting} problems, as these are the ones that we as humans care about (and thus put into the test set). Formalizing a notion of interestingness is the holy grail of task generation, but has remained elusive thas far (\citet{schmidhuber2009simple} makes an attempt based on compression).

However, there may be a solution to the interestingness problem in theorem proving.
The tasks in theorem proving are conjectures; once proven, they become theorems which themselves may be used in the proofs of other conjectures. This provides concrete feedback about which tasks (theorems) are in fact \emph{useful}, arguably an important criterion for what interests human mathematicians. Once quantified, this notion of usefulness could be learned and optimized for by a proposer.

%% file: appendix.tex
\section{Axiom Sets}

\begin{table}[h]
    \centering
    \begin{tabular}{cccc}
    \toprule
    {Axiom Set} &                  {Domain} &  {Axioms} &  {Theorems} \\
    \midrule
           FLD1 &              Field Theory &        27 &          78 \\
           FLD2 &              Field Theory &        26 &         105 \\
           GEO6 &                  Geometry &        46 &         128 \\
           GEO7 &                  Geometry &        58 &          38 \\
           GEO8 &                  Geometry &        35 &         128 \\
           GEO9 &                  Geometry &        66 &          37 \\
           GRP5 &              Group Theory &         7 &          10 \\
           KRS1 &  Knowledge Representation &       108 &          41 \\
           NUM9 &             Number Theory &        42 &          30 \\
           SET1 &                Set Theory &        24 &          11 \\
    \bottomrule
    \end{tabular}
    \caption{\label{table:datasets}Axiom sets and human-written theorems extracted from TPTP.}
\end{table}

\section{Forward Proposer}

\subsection{Forward Proposer Settings}
\label{appendix:fwdp-settings}

See Table~\ref{table:fwdp-settings} for the numbers of forward steps and temperatures used in the forward proposer for each axiom set.

\begin{table}[h]
    \centering
    \begin{tabular}{ccc}
    \toprule
    {Axiom Set} & {Forward Steps} &  {Temperature} \\
    \midrule
           FLD1 &        15 &          10 \\
           FLD2 &        10 &          12 \\
           GEO6 &        10 &           8 \\
           GEO7 &         5 &          12 \\
           GEO8 &        10 &           8 \\
           GEO9 &        15 &           4 \\
           GRP5 &        10 &          20 \\
           KRS1 &         5 &    $\infty$ \\
           NUM9 &        10 &          20 \\
           SET1 &        10 &           5 \\
    \bottomrule
    \end{tabular}
    \caption{Forward proposer settings.}
    \label{table:fwdp-settings}
\end{table}

\subsection{Limitations of the Forward Proposer}
\label{appendix:fwdp-limitations}

It should also be noted that, although the resolution calculus is implication-complete~\citep{lee67} (\emph{i.e} the forward proposer is capable of generating (almost) any clause implied by the axioms)~\footnote{"Lee's theorem" states that for any clause $C$ that is implied by the axioms, there is a clause $C'$ that can be generated by the resolution calculus such that $C'$ \emph{subsumes} $C$. For more details, see \citet{fitting12}.}, there is a class of theorems that it cannot generate: those whose conclusion cannot be expressed as a single clause. This includes theorems that introduce new symbols, new axioms and hypotheses, or existential quantifiers. However, most TPTP theorems (77.7\%) are in fact forward-proposable; see Appendix~\ref{appendix:forward-proposability}.

\subsection{Forward Proposability of TPTP Theorems}
\label{appendix:forward-proposability}

A theorem is considered forward-proposable with respect to its axiom set if it can be written in the form Axioms $\rightarrow$ Conjecture Clause. Whether this is the case may not be immediately obvious from a TPTP problem statement; for example, many TPTP problems are already negated and CNF-ized for refutation proving. However, if a theorem is in fact forward-proposable, its negated CNF will contain two clause sets: the axioms, and the (unit) clauses corresponding to the negation of the conjecture clause. Thus, it suffices to construct a clause whose functors and predicates appear in the axioms, and whose negation produces the non-axiom clauses (given appropriate namings of Skolem constants). This is possible exactly when each non-axiom clause a) is unit, b) contains no variables, and c) only uses symbols from the axioms, with the exception of constants (which are Skolemized variables). Table~\ref{table:forward-proposability} summarizes the forward-proposability of TPTP problems across the various domains.

\begin{table}
    \centering
    \begin{tabular}{cccc}
    \toprule
    {Axiom Set} & {Forward-Proposable} & {Total} & {Percent} \\
    \midrule
    FLD1 & 75 & 78 & 96.2 \\
    FLD2 & 102 & 105 & 97.1 \\
    GEO6 & 95 & 128 & 74.2 \\
    GEO7 & 33 & 38 & 86.8 \\
    GEO8 & 69 & 128 & 53.9 \\
    GEO9 & 33 & 37 & 89.2 \\
    GRP5 & 10 & 10 & 100.0 \\
    KRS1 & 40 & 41 & 97.6 \\
    NUM9 & 4 & 30 & 13.3 \\
    SET1 & 10 & 11 & 90.9 \\
    TOTAL & 471 & 606 & 77.7 \\
    \bottomrule
    \end{tabular}
    \caption{\label{table:forward-proposability}Forward-proposability by axiom set.}
\end{table}

\section{Details of the resolution prover}
\label{appendix:saturation}
In this section, we describe the details of our resolution prover. The pseudo-code can be found in Algorithm~\ref{alg:saturation}. The procedures $\theta$\code{-subsumption}, \code{find_resolutions}, and \code{find_factors} are the same as for other provers~\citep{riazanov&voronkov02,schulz&al19}. 

The main procedure is \code{refute} and takes three inputs: the initial set of clauses (this includes axioms and negated conjecture clauses), a cost function\footnote{This can be a handcrafted heuristic or computed by a neural network.} which takes as input a clause and outputs its cost, and the age-cost ratio hyperparameter $a:c$. It maintains two priority queues at any given time: an age priority queue \code{qa} and a cost priority queue \code{qc}. The age priority queue is ordered solely by the iteration number at which a clause is generated, which ensures that every generated clause is processed after a finite number of iterations. The cost priority queue is ordered by the output of the cost function. The algorithm also maintains a set of processed clauses $P$. To begin with, all the initial clauses are inserted into both priority queues.

At each iteration of the algorithm, first we select a priority queue based on the age-cost ratio $a:c$: The age queue is selected for $a$ consecutive iterations, then the cost queue is selected for $c$ consecutive iterations an so on.
After selecting a queue, we select the clause $C_t$ that is at the top of this queue and remove it from both queues. If the clause is the empty clause, a refutation has been found and the theorem is proved.
Otherwise, the algorithm then conducts standard subsumption checks for the selected clause $C_t$ with the existing set of processed clauses $P$ (initially empty). Specifically, we check  forward and backward $\theta$-subsumption~\citep{plotkin1970note} to remove unnecessary clauses. Forward subsumption checks if the selected clause $C_t$ is less general than any clause in the processed set. 
A clause $C_1$ $\theta$-subsumes a clause $C_2$ if there exists a substitution $\theta$ that when
applied to $C_1$ gives $C_2$.
Circularity of subsumption checks is avoided by performing forward subsumption before backward subsumption.
To avoid another pitfall of subsumption checks where resolution can produce
clauses that increase in size but also in generality at the same time,
subsumed clauses are removed only if they pass an additional test: we say that a clause $C_1$ \emph{order-subsumes} a clause $C_2$
if $C_1$ has no more literals as $C_2$ and if $C_1$ $\theta$-subsumes $C_2$.
If the selected clause is subsumed by any existing clause in the processed set, 
it is simply discarded and we proceed to the next iteration with the appropriate age or cost queue. 
Otherwise, we proceed to check for backward subsumption: if any clause in the processed set $P$ is order-subsumed by $C_t$, it is removed from $P$.
Then, we compute all possible inferences (resolutions and factors) of $C_t$ with the remaining clauses in the processed set $P$, and the generated clauses are inserted in the age and cost priority queues. We also insert the clause $C_t$ in the processed set. The algorithm iterates until the queues are empty, which indicates that a refutation cannot be found and the initial set of clauses is satisfiable, meaning the theorem is not true. 

Observe that since all the initial clauses (axioms, hypotheses, etc.) are initially inserted in the priority queues, they are be subject to subsumption checks when selected. We also do simple syntactic tautology elimination. This is done by matching the negative literals of the clause to the positive literals of the clause syntactically. If there is a one to one match, the clause is marked as tautology and eliminated at the time of generation.

\newcommand{\activeset}{\text{active}}

\begin{algorithm}
\begin{lstlisting}
def order_subsumes($C_1$, $C_2$):
  return num_literals($C_1$) $\leq$ num_literals($C_2$) and $\theta$-subsumes($C_1$, $C_2$)

def refute(initial_clauses, cost_fn, age_cost_ratio):
  a = numerator(age_cost_ratio)      # a in the ratio a:c
  c = denominator(age_cost_ratio)    # c in the ratio a:c
  qa = make_priority_queue(age)      # age queue of unprocessed clauses
  qc = make_priority_queue(cost_fn)  # cost queue of unprocessed clauses
  $P$ = {}                              # set of processed clauses
  qa.insert(initial_clauses)
  qc.insert(initial_clauses)
  t = 0
  while not qc.empty():
    if t % (a + c) < a:
      # Select the oldest unprocessed clause.
      $C_t$ = qa.extract_min()
      qc.remove($C_t$)
    else:
      # Select the unprocessed clause with the least cost.
      $C_t$ = qc.extract_min()
      qa.remove($C_t$)

    if is_empty_clause($C_t$):
      return "refutation_found"  # i.e. unsatisfiable

    # FORWARD SUBSUMPTION
    # Discard $C_t$ if it is order-subsumed by a clause in $P$.
    if $\exists C\in P$ s.t. order_subsumes($C$, $C_t$):
      continue

    # BACKWARD SUBSUMPTION
    # Discard any clause in $P$ that is order-subsumed by $C_t$.
    for $C$ in $P$:
      if order_subsumes($C_t$, $C$):
        $P$ = $P\setminus\{C\}$
  
    # Enqueue the factors of $C_t$ and any resolutions between $C_t$ and the clauses in $P$.
    new_clauses = find_factors($C_t$) $\cup$ find_resolutions($C_t$, $P$)
    qa.insert(new_clauses)
    qc.insert(new_clauses)

    # Add $C_t$ to the set of processed clauses.
    $P$ = $P\cup\{C_t\}$
    
    t = t + 1

  return "refutation_not_found"  # i.e. satisfiable
\end{lstlisting}
\caption{The saturation algorithm.}
\label{alg:saturation}
\end{algorithm}

\section{Feature Representations and Neural Architectures}
\label{appendix:architectures}

Here we describe in more detail the neural architectures and the inputs for those architectures. All neural networks were written in the Jax framework \citep{jax2018github}.

\subsection{Multi-Layer Perceptron}

We used seven scalar features to represent each clause: number of negated literals, number of positive literals, number of atomic terms, number of distinct predicates, number of distinct functors, number of distinct variables and total number of variables. The features of the initial clauses were further aggregated via four different aggregation functions: sum, average, maximum and minimum. Finally, three more scalars were concatenated to these: the number of the step at which the clause to be evaluated was generated, the number of premises used in the inference that generated the clause (between 0 and 2 depending on the type of inference used) and the number of initial clauses. In total, there were 38 elements in the input vector.

We opted for a five layer MLP with layer sizes 256, 64, 16, 4 and 1, and ReLU activation \citep{nair2010rectified}. In training, the batch size was fixed to 4096 and learning rate to $10^{-4}$.

\subsection{Transformer}

We used a vanilla 6-layer transformer architecture with a hidden size of 128 per node, a hidden size of 256 in the residual blocks, and 8 attention heads. The graphical inputs were capped to 256 nodes by dropping any additional nodes past the first 256. Symbols were hashed to 16-dimensional vectors following a standard normal distribution. The batch size was fixed to 256 and the learning rate to $10^{-4}$.`

\subsection{Cost Function}

We compute the cost of a clause using a mixture of the clause's (negative) probability of appearing the proof, as predicted by the neural network, and the clause's size (number of symbols), with a ratio of 64:1. This means that early on in training, when the neural net cannot yet distinguish between its different inputs, the cost function falls back on the clause size, as in the ``basic'' prover.

\section{Experimental Details}
\label{appendix:experiments}

Our experiments were distributed across many machines and used several types of workers working in concert:

\begin{enumerate}
    \item A \textbf{replay buffer} which stores the positive and negative examples. The buffer had a maximum size of 65536, and examples were sampled and removed uniformly at random. We used the open-source Reverb implementation \citep{Reverb}.
    \item A \textbf{learner} which trains the neural network on data sampled from the replay buffer. In our experiments the learner ran on a V100 GPU.
    \item Many \textbf{actors} which repeatedly sample problems from the forward proposer and attempt to solve them. Solved problems result in positive and negative clause samples which are placed into the replay buffer as described in Section~\ref{section:prover}. Periodically (every minute), each actor pulls the latest version of the network parameters from the learner.
\end{enumerate}

During training, actors are given a 5-minute time limit per problem. The MLP experiments used 256 actors running on CPUs, while the transformer experiments used 1024 actors sharing a pool of 32 Google V2 TPUs for inference (MLPs were not found to benefit from TPU acceleration). All reported results were taken after two days of training. This may seem to unfairly disadvantage the MLPs computationally, but letting the MLPs run for an additional 4 days saw no improvements on any axiom set.